**SAR image segmentation algorithms based on I-divergence-TV model**

Guangming liu
Yantai Science and Technology Association,email:1733823365@qq.com

**Abstract**— In this paper, we propose a novel variational active contour model based on I-divergence-TV model to segment Synthetic aperture radar (SAR) images with multiplicative gamma noise, which hybrides edge-based model with region-based model. The proposed model can efficiently stop the contours at weak or blurred edges, and can automatically detect the exterior and interior boundaries of images. We further transform the proposed model into a general ROF model by adding a proximity term ,and it can be solved by a fast denoising algorithm proposed by Jia-Zhao or soved by BM3D and NLM denoising algorithm, which also provide a unified solution framework for formally generalized-ROF-like subproblems arising in multivariate splitting algorithms[25]. [25] was submitted on 29-Aug-2013, and our early edition was ever submitted to TGRS on 12-Jun-2012, Venkatakrishnan et al. [26] proposed their **"PnP algorithm"** on 29-May-2013, so Venkatakrishnan and we proposed the **"PnP algorithm"** almost simultaneously. Experimental results for synthetic images and real SAR images show that the proposed fast fixed point algorithm is robust and efficient compared with the state-of-the-art approach.

## Ⅰ.INTRODUCTION

Sea /land segmentation in satellite-borne synthetic aperture radar (SAR) images is a key step in coastline extraction and elsewhere, which is with only two classes to assign.

It is well known that SAR images segmentation is usually recognized as a complex problem, which plays a fundamental role in the SAR images interpretation. SAR images are affected by speckle, a multiplicative noise that gives the images a grainy appearance and makes the interpretation of SAR images a more difficult task to achieve. Image segmentation technique in airborne or satellite-borne SAR images is a powerful tool in environmental monitoring applications such as coastline extraction, object detection and so forth.

Markov random fields[1-2], present many interesting properties. Indeed, they not only allow one to design segmentation techniques which are able to take into account the nature of the speckle noise in a statistically optimal way but they also provide an efficient regularization method. However, Markov random field models introduce ad hoc parameters which can not be easily automatically determined and which can lead to difficult optimization problem.

A polygonal snakes[3-4] is parameter free. Also it does not assume the number of regions known, whereas current level set methods do. However, they have significant limitations which level sets remove: they cannot segment regions of arbitrary topology. In particular, they cannot segment regions composed of disjoint parts. Snakes do not allow changes of active curve topology during evolution.

Level set methods (LSM) have been extensively applied to SAR images segmentation [5]-[8]. We know that level set methods can be divided into two major categories: region-based models and edge-based models. In order to make use of edge information and region characteristic, region-based model is often integrated with edge-based model to form the energy functional of models. There are some advantages of level set methods over classical image segmentation methods, such as edge detection, thresholding, and region grow: level set methods can give sub-pixel accuracy of object boundaries, and can be easily formulated under a energy functional minimization, and can provide smooth and closed contour as a result of segmentation and so forth.

However, a common difficulty with above variational image segmentation models [5-10] is that the energy functionals to be minimized has local minima (which are not global minima). This is a much more serious draw back than nonuniqueness of global minimizers (which is also a common phenomenon) because local minima of segmentation models often have completely wrong levels of detail and scale: whereas global minimizers of a given model are usually all reasonable solutions, the local minima tend to be blatantly false. Many solution techniques for variational models are based on gradient descent, and are therefore prone to getting stuck in such local minima.

In order to resolve the problems associated with non-convex models, Chan-Esedoglu-Nikolova [11] proposed a convex relaxation approach for image segmentation models. But the global convex segmentation models contain a total variation (TV) term that is not differentiable in zero, making them difficult to compute. Chan etal. [11] proposed to either enforce the inequality constraint using an exact penalty function, which is nondifferentiable, or regularize the penalty function, which does not exactly enforce the inequality constraint. Bresson etal. [12] used a splitting/regularization approach to minimize(1). Their method "smears" the values of u near the object boundaries, and thus makes the segmentation results more dependent on the cut-off parameter T, which could eliminate the segmentation details.

It is well known that the split Bregman method is a technique for fast minimization of $l^1$ regularized energy functional. Goldstein-Osher (GO) [13-14] proposed a more efficient way to compute the TV term by applying the split Bregman method. But GO algorithm still requires solving a partial difference equation in each iteration step.

Jia-Zhao[15-16] give a algorithm for the solution of ROF denoising model [17] based on anisotropic TV term, which is much faster than GO algorithm.

Recently, Steidl and Teuber [18] examined theoretically and numerically the suitability of the I-divergence-TV convex denoising model for restoring images contaminated by multiplicative Gamma noise, which is the typical data fitting term when dealing with Poisson noise. In this paper, we propose a novel variational active contour model based on I-divergence-TV model for SAR image segmentation. The novel model can address well non-texture speckle image segmentation .On the other hand, it is well known that the ROF denoising model is strictly convex, which always admits a unique solution. We can transform the proposed model into classic ROF model by adding a proximity term. We further give a fast fixed point algorithm to solve the global convex segmentation question based on proposed novel model, which do not involve partial differential equations.

This paper is organized as follows. Section Ⅱ describes the proposed variational active contour model. Section Ⅲ describes the proposed two fast globlly convex segmentation algorithms, and Section Ⅳ describes the experimental results. Section Ⅴ concludes this paper.

## Ⅱ. NOVEL VARIATIONAL ATIVE CONTOUR MODEL BASED ON I-DIVERGENCE-TV MODEL

We denote by $R^2$ the usual 2-dimensional Euclidean space $H$. We use $\langle .,. \rangle$ and $\| \|_2$, respectively, to denote the inner product and the corresponding $l^2$ norm of an Euclidean space $H$ while $\| \|_1$ is used to denote the $l^1$ norm. We denote by $\nabla_x^T$ ($\nabla_y^T$) the conjugate of the gradient operator $\nabla_x$ ($\nabla_y$).

Given a observed intensity image $f : \Omega \rightarrow R (f > 0)$, where $\Omega$ is a bounded open subset of $R^2$, speckle is well modeled as a multiplicative random noise $n$, which is independent of the true image $u$, i.e. $f = u \cdot n$. We know that fully developed multiplicative speckle noise is Gamma distributed with mean value $\mu_n = 1$ and variance $\sigma_n{}^2 = 1/L$, where $L$ is the equivalent number of independent looks of the image. Steidl and Teuber (ST) [18] examined theoretically and numerically the suitability of the I-divergence-TV convex denoising model for restoring images contaminated by multiplicative Gamma noise, which is the typical data fitting term when dealing with Poisson noise.i.e.

$$E(u) = \int_{\Omega} |\nabla u| dx + \mu \int_{\Omega} (u - f \log u) dx \ . \qquad (1)$$

The first term of (1) is the TV regularization term, and the second term is the data fitting term, and $\mu$ is a positive constant parameter to balance the first term and second term. The distinctive feature of the TV regularization term and its various variants is that edges of images are preserved in the denoised image.

We adopt the idea of Chan-Vese (CV) model [9] on the two-phase segmentation question, and suppose that the true image $u$ is piecewise constant, i.e. $u = C_1$ for $x \in \Omega_1$, and $u = C_2$ for $x \in \Omega_2$, where $\{\Omega_1, \Omega_2\}$ is a partitioning of the image domain $\Omega$. Equation (1) can be written in the level set formulation by adopting the idea of the CV model as:

$$\begin{aligned} E(\phi, C_1, C_2) &= \int_{\Omega} |\nabla H(\phi)| dx + \mu \cdot [\int_{\Omega} (C_1 - f \log C_1) \cdot H(\phi) dx \\ &+ \int_{\Omega} (C_2 - f \log C_2)(1 - H(\phi)) dx] \end{aligned} \qquad (2)$$

Where $\phi$ is a level set function; $H(\phi)$ is the Heaviside function, which is approximated by a smooth function $H_{\varepsilon}(\phi) = \frac{1}{2}[1 + \frac{2}{\pi} \arctan(\frac{\phi}{\varepsilon})]$ to automatically detect interior contours and insure the computation of a global minimizer in this paper. The derivative of $H_{\varepsilon}(\phi)$ is also approximated by a smooth function $\delta_{\varepsilon}(\phi) = H_{\varepsilon}^{'}(\phi) = \frac{1}{\pi}(\frac{\varepsilon}{\varepsilon^2 + \phi^2})$. Using the level set formulation, the true image $u$ can be expressed:

$$u = C_1 H(\phi) + C_2 (1 - H(\phi)) \qquad (3)$$

Therefore, the energy functional (2) becomes:

$$E_{\varepsilon}(\phi, C_1, C_2) = \int_{\Omega} \delta_{\varepsilon}(\phi) |\nabla \phi| dx + \mu \cdot [\int_{\Omega} (C_1 - f \log C_1) \cdot H_{\varepsilon}(\phi) dx \\ + \int_{\Omega} (C_2 - f \log C_2) \cdot (1 - H_{\varepsilon}(\phi)) dx] \quad (4)$$

We further modify equation (4) to incorporate information from an edge detector $g$, which can make (4) more likely to favor segmentation along curves where the edge detector function $g$ is minima. On the other hand, in order to penalize the deviation of $\phi$ from a signed distance function during its evolution, we add a level set regularization term [10] to equation (4). Thus we propose a novel variational active contour model based on I-divergence-TV denoising model, which hybrides edge-based model with region-based model, named as GID, i.e.

$$E_{\varepsilon,g}(\phi, C_1, C_2) = \int_{\Omega} g \delta_{\varepsilon}(\phi(x)) |\nabla \phi(x)| dx + \mu \cdot [\int_{\Omega} (C_1 - f(x) \log C_1) \cdot H_{\varepsilon}(\phi(x)) dx \\ + \int_{\Omega} (C_2 - f(x) \log C_2) \cdot (1 - H_{\varepsilon}(\phi(x)) dx] + v \cdot \int_{\Omega} \frac{1}{2} (|\nabla \phi(x)| - 1)^2 dx \quad (5)$$

Where $g$ is a positive and decreasing edge detector function which is often defined as $g = \dfrac{1}{1 + \beta |\nabla f_{\sigma} \otimes I|^2}$, and $g$ usually takes smaller values at object boundaries than at other locations; The parameter $\beta$ controls the details of the segmentation, and $\nabla f_{\sigma} \otimes I$ is used to smooth the image to reduce the noise; In order to efficiently smooth multiplicative noise, we use the infinite symmetric exponential filter (ISEF) $f_{\sigma}(x) = \dfrac{1}{2\sigma} e^{-\frac{|x|}{\sigma}}$ with standard deviation $\sigma = 1.2$ and of size $15 \times 15$ in edge detector $g$, which is optimal in the case of multiplicative noise [19]; And ISEF has better edge localization precision than other edge detectors [20].

For fixed level set function $\phi$, we minimize the function $E_{\varepsilon,g}(\phi, C_1, C_2)$ with respect to the constant $C_1$ and $C_2$. By calculus of variations, it is easy to solve them

by

$$C_1 = \frac{\int_\Omega f H_\varepsilon(\phi)dx}{\int_\Omega H_\varepsilon(\phi)dx}, \quad C_2 = \frac{\int_\Omega f(1-H_\varepsilon(\phi))dx}{\int_\Omega (1-H_\varepsilon(\phi))dx}. \tag{6}$$

For fixed $C_1$ and $C_2$, the level set function $\phi$ that minimizes $E_{\varepsilon,g}(\phi, C_1, C_2)$ can be obtained by solving the gradient flow equation:

$$\frac{\partial \phi}{\partial t} = \delta_\varepsilon(\phi) div(g\frac{\nabla \phi}{|\nabla \phi|}) - \mu \cdot \delta_\varepsilon(\phi) \cdot \eta + v \cdot (\nabla^2 \phi - div(\frac{\nabla \phi}{|\nabla \phi|})) \tag{7}$$

Where $\eta = (C_1 - f \log C_1) - (C_2 - f \log C_2)$ .

If we replace

$\eta = (C_1 - f \log C_1) - (C_2 - f \log C_2)$ with $\eta = (\log C_1 + f/C_1) - (\log C_2 + f/C_2)$ , and remove the edge detector function $g$, then equation (7) is just the model proposed by Ayed et al.[8].

The first term has a smoothing effect on the zero level contour, which is necessary to maintain the regularity of the contour. The second term is referred to as the data fitting term, this term plays a key role in the proposed model, since it is responsible for driving the active contour to ward object boundaries. The third term is called a level set regularization term, since it serves to maintain the regularity of the level set function.

Because of the diffusion term introduced by our level set regularization term, we no longer need the upwind scheme as in the traditional level set methods. Instead, all the spatial partial derivatives $\frac{\partial \phi}{\partial x}$ and $\frac{\partial \phi}{\partial y}$ in (7) can be simply discretized as central finite differences.The temporal partial derivative $\frac{\partial \phi}{\partial t}$ is discretized as a forward difference.

## III. PROPOSED FIXED POINT ALGORITHM

### A. Global Convex Segmentation Model

It is well known that energy functional (5) may not have a unique global minimizer because it is non-convex. In this section, we introduce a new GID model, which incorporates the global convex segmentation (GCS) method [11].

Considering the gradient flow equation (7), we first drop the last term which regularized the level set function to be close to a distance function, then we obtain a new gradient flow equation as:

$$\frac{\partial \phi}{\partial t} = \delta_\varepsilon(\phi)[div(g\frac{\nabla \phi}{|\nabla \phi|}) - \mu \cdot \eta] \tag{8}$$

Where $\eta = (C_1 - f\log C_1) - (C_2 - f\log C_2)$ or $\eta = (\log C_1 + f/C_1) - (\log C_2 + f/C_2)$ .

Following the idea in Chan et al. [11], the stationary solution of gradient flow equation (8) coincides with the stationary solution of the simplied flow:

$$\frac{\partial \phi}{\partial t} = div\left(g\frac{\nabla \phi}{|\nabla \phi|}\right) - \mu \cdot \eta \tag{9}$$

We now propose a new energy functional as follows:

$$E(\phi) = \int_\Omega g\,|\nabla \phi|dx + \mu\int_\Omega \eta \cdot \phi dx \tag{10}$$

It can be clearly seen that the simplied flow (9) is just the gradient descent flow of energy functional (10). Thus the minimization problem we want to solve is

$$\min E(\phi) = \min \int_\Omega g\,|\nabla \phi|dx + \mu\int_\Omega \eta \cdot \phi dx \tag{11}$$

To guarantee the global minimum, we must constrain the solution to lie in a finite interval. In this paper, we restrict the solution $\phi$ to lie in a finite interval ( $0 \le \phi \le 1$ ) as follows:

$$\min_{0\le\phi\le 1} E(\phi) = \min_{0\le\phi\le 1} \int_\Omega g\,|\nabla \phi|dx + \mu\int_\Omega \eta \cdot \phi dx \tag{12}$$

The segmented region can be found by thresholding the function $\phi(x)$ for some $0<\gamma<1$: $\Omega_1=\{x:\phi(x)>\gamma\}$.

The first term in the proposed energy functional (10) is total variation (TV) norm:

$$TV(\phi)=\int g\left|\nabla\phi\right|dx=\left\|\nabla\phi\right\|_g \tag{13}$$

Thus the proposed minimization problem becomes

$$\min_{0\le\phi\le1} E(\phi)=\min_{0\le\phi\le1}\left(\left\|\nabla\phi\right\|_g+\mu<\phi,\eta>\right) \tag{14}$$

**B. Split Bregman method**

It is well known that the split Bregman method is a technique for fast minimization of $l^1$ regularized energy functional. A break through was made by Goldstein-Osherin (GO) [13-14].They proposed a more efficient way to compute the TV term based on the split Bregman method. To apply the split Bregman method to (14), we introduce auxiliary variables $d_x\leftarrow\nabla_x\phi$, $d_y\leftarrow\nabla_y\phi$, and add a quadratic penalty function to weakly enforce the resulting equality constraint which results in the following unconstrained problem:

$$\begin{aligned}&(\phi^{k+1},d_x^{k+1},d_y^{k+1})=\arg\min_{0\le\phi\le1}\left\|d_x\right\|_g+\left\|d_y\right\|_g+\mu<\phi,\eta^k>+\\&\frac{\lambda}{2}\left\|d_x-\nabla_x\phi\right\|^2+\frac{\lambda}{2}\left\|d_y-\nabla_y\phi\right\|^2\end{aligned} \tag{15}$$

We then apply split Bregman method to strictly enforce the constraints $d_x=\nabla_x\phi$, $d_y=\nabla_y\phi$. The resulting optimization problem becomes:

$$\begin{aligned}&(\phi^{k+1},d_x^{k+1},d_y^{k+1})=\arg\min_{0\le\phi\le1}\left\|d_x\right\|_g+\left\|d_y\right\|_g+\mu<\phi,\eta^k>+\\&\frac{\lambda}{2}\left\|d_x-\nabla_x\phi-b_x^k\right\|_2^2+\frac{\lambda}{2}\left\|d_y-\nabla_y\phi-b_y^k\right\|_2^2\\&b_x^{k+1}=b_x^k+\nabla_x\phi^{k+1}-d_x^{k+1}\\&b_y^{k+1}=b_x^k+\nabla_y\phi^{k+1}-d_y^{k+1}\end{aligned} \tag{16}$$

For fixed $\vec{d}$, the Euler-Lagrange equation of optimization problem (16) with

respect to $\phi$ is:

$$\Delta\phi^{k+1} = \frac{\mu\cdot\eta^k}{\lambda} - \nabla_x^T(d_x^k - b_x^k) - \nabla_y^T(d_y^k - b_y^k) \tag{17}$$

For fixed $\phi$, minimization of (16) with respect to $\vec{d}$ gives:

$$\begin{aligned} d_x^{k+1} &= shrink_{g/\lambda}(\nabla_x\phi^{k+1} + b_x^k) \\ d_y^{k+1} &= shrink_{g/\lambda}(\nabla_x\phi^{k+1} + b_y^k) \end{aligned} \tag{18}$$

Where $shink_{g/\lambda}(x) = \mathrm{sgn}(x)\max(|x| - g/\lambda, 0)$.

By using central discretization for Laplace operator and backward difference for divergence operator, the numerical scheme for (17) becomes:

$$\begin{aligned} \alpha_{i,j} &= d_{i-1,j}^x - d_{i,j}^x - b_{i-1,j}^x + b_{i,j}^x + d_{i,j-1}^y - d_{i,j}^y - b_{i,j-1}^y + b_{i,j}^y \\ \beta_{i,j} &= \frac{1}{4}\left(\phi_{i-1,j} + \phi_{i+1,j} + \phi_{i,j-1} + \phi_{i,j+1} - \frac{\mu\cdot\eta^k}{\lambda} + \alpha_{i,j}\right) \\ \phi_{i,j} &= \max(\min(\beta_{i,j}, 1), 0) \end{aligned} \tag{19}$$

As the optimal $\phi$ is found, the segmented region can be found by thresholding the level set function $\phi(x)$ for some $\gamma \in (0,1)$ : $\Omega_1 = \{x : \phi(x) > \gamma\}$. The split Bregman algorithm for the minimization problem (14) can be summarized as follows:
split Bregman method for anisotropic TV –based Model

Given: noisy image $f$ ; $\lambda > 0$ , $\mu > 0$
Initialization: $b^0 = 0$ , $d^0 = 0$ , $\phi^0 = f/\max(f)$
For $k = 0,1,2,\cdots$
compute(20)
$d_x^{k+1} = shrink_{g/\lambda}(\nabla_x\phi^{k+1} + b_x^k)$
$d_y^{k+1} = shrink_{g/\lambda}(\nabla_x\phi^{k+1} + b_y^k)$
$b_x^{k+1} = b_x^k + \nabla_x\phi^{k+1} - d_x^{k+1}$
$b_y^{k+1} = b_x^k + \nabla_y\phi^{k+1} - d_y^{k+1}$
END

**C. Proposed Fixed Point Algorithm (FPA)**

We know the ROF denoising model always admits a unique solution because the energy functional is strictly convex. In order to make sure that equation (15) is strictly convex, we reformulate equation (14) by adding the proximal term $\frac{\alpha}{2} \cdot \left\| \phi - \phi^k \right\|_2^2$ with $\alpha > 0$ . Supposing the weighted TV term (13) in minimization problem (14) is anisotropic, then the minimization problem (14) becomes:

$$
\begin{aligned}
\phi^{k+1} &= \arg\min_{0\le\phi\le1} \left\| \nabla_x \phi \right\|_g + \left\| \nabla_y \phi \right\|_g + \mu < \phi, \eta^k > + \frac{\alpha}{2} \left\| \phi - \phi^k \right\|_2^2 \\
&= \arg\min_{0\le\phi\le1} \left\| \nabla_x \phi \right\|_g + \left\| \nabla_y \phi \right\|_g + \mu < \phi - \phi^k, \eta^k > + \frac{\alpha}{2} \left\| \phi - \phi^k \right\|_2^2 \\
&= \arg\min_{0\le\phi\le1} \left\| \nabla_x \phi \right\|_g + \left\| \nabla_y \phi \right\|_g + \frac{\alpha}{2} \left\| \phi - (\phi^k - \frac{\mu\eta^k}{\alpha}) \right\|_2^2
\end{aligned}
\tag{20}
$$

Treating $\phi^k - \frac{\mu\eta^k}{\alpha}$ **as the "noisy image",** (20) minimizes the residue between the **"clean image"** $\phi$ and $\phi^k - \frac{\mu\eta^k}{\alpha}$ using the total variation norm, then (20) is the total variation denoising problem. In the past, solutions of the TV model were based on nonlinear partial differential equations and the resulting algorithms were very complicated. We propose a fast fixed point algorithm (FPA) to solve (20) based on JZ algorithm [15] as follows:

$$
\begin{aligned}
& b_x^k = (I - shrink_{g/\lambda})(\nabla_x \phi^k + b_x^{k-1}) \\
& b_y^k = (I - shrink_{g/\lambda})(\nabla_y \phi^k + b_y^{k-1}) \\
& \phi^{k+1} = \phi^k - \frac{\mu\eta^k}{\alpha} - \frac{\lambda}{\alpha}(\nabla_x^T b_x^k + \nabla_y^T b_y^k) \\
& \phi^{k+1} = \max(\min(\phi^{k+1}, 1), 0)
\end{aligned}
\tag{21}
$$

Where $shink_{g/\lambda}(x) = \operatorname{sgn}(x)\max(|x| - g/\lambda, 0)$ .

We can see that the proposed FPA algorithm is very simple and does not involve partial differential equations or difference equation.

In order to further accelerate the convergence of (21) , we adopt the following iteration scheme by utilizing k-averaged operator theory ( see more details in [21] ):

$$
\begin{aligned}
b_x^k &= t\cdot b_x^k + (1-t)\cdot (I - shrink_{g/\lambda})(\nabla_x \phi^k + b_x^{k-1}) \\
&= t\cdot b_x^{k-1} + (1-t)\cdot (I - shrink_{g/\lambda})[\nabla_x(\phi^{k-1} - \frac{\mu\eta^{k-1}}{\alpha} - \frac{\lambda}{\alpha}\nabla_y^T b_y^{k-1}) + (I - \frac{\lambda}{\alpha}\nabla_x\nabla_x^T) b_x^{k-1}] \\
b_y^k &= t\cdot b_y^k + (1-t)\cdot (I - shrink_{g/\lambda})(\nabla_y \phi^k + b_y^{k-1}) \\
&= t\cdot b_y^{k-1} + (1-t)\cdot (I - shrink_{g/\lambda})[\nabla_y(\phi^{k-1} - \frac{\mu\eta^{k-1}}{\alpha} - \frac{\lambda}{\alpha}\nabla_x^T b_x^{k-1}) + (I - \frac{\lambda}{\alpha}\nabla_y\nabla_y^T) b_y^{k-1}]
\end{aligned}
\tag{22}
$$

Where the weight factor $t \in (0,1)$ is called the relaxation parameter.

We can easily infer that operator $(I - shrink_{g/\lambda})(I - \frac{\lambda}{\alpha}\nabla_x\nabla_x^T)$ and $(I - shrink_{g/\lambda})(I - \frac{\lambda}{\alpha}\nabla_y\nabla_y^T)$ in (18) are all nonexpansive only if $\frac{\lambda}{\alpha}$ is less than $\frac{1}{4}\sin^{-2}\frac{(N-1)\pi}{2N}$ which is slightly bigger than 1/4 [21-22]. According to the classic fixed theory, the sequences of $b_x^k$ and $b_y^k$ can converge to fixed points of them.

The FPA for the minimization problem (21) can be summarized as follows:

FPA for anisotropic TV–based Model:

Given: noisy image $f$ ; $\lambda > 0$ , $\mu > 0$ , $\alpha > 0$ , $t \in (0,1)$ , $\gamma \in (0,1)$

Initialization: $b_x^0 = 0$ , $b_y^0 = 0$ , $\phi^0 = f/\max(f)$ , $\Omega_0 = \{x : \phi^0 > \gamma\}$ ,he

$$C_1^0 = \int_{\Omega_0} f dx ,\quad C_2^0 = \int_{\Omega_0^c} f dx$$

For $k = 0,1,2,\cdots$

$$b_x^k = t\cdot b_x^{k-1} + (1-t)\cdot (I - shrink_{g/\lambda})(\nabla_x \phi^k + b_x^{k-1})$$

$$b_y^k = t\cdot b_y^{k-1} + (1-t)\cdot (I - shrink_{g/\lambda})(\nabla_y \phi^k + b_y^{k-1})$$

$$\phi^{k+1} = \phi^k - \frac{\mu\eta^k}{\alpha} - \frac{\lambda}{\alpha}(\nabla_x^T b_x^k + \nabla_y^T b_y^k)$$

$$\phi^{k+1} = \max(\min(\phi^{k+1},1),0)$$

$$\Omega_{k+1} = \{x : \phi(x) > \gamma\} ,\quad C_1^{k+1} = \int_{\Omega_{k+1}} f dx , C_2^{k+1} = \int_{\Omega_{k+1}^c} f dx$$

END

**D. Quantitative Evaluation:**

We give the following several measures to evaluate the proposed FPA

algorithm in this paper .

**1) Uniformity measurement**

We first adopt the uniformity measurement of image segmentation regions to evaluate the performance of the proposed method. The interior of each region should be uniform after the segmentation and there should be a great difference among different regions. That is to say, the uniformity degree of regions represents the quality of the segmentation. Therefore, we give the measurement of segmentation accuracy (SA) [23] as:

$$pp = 1 - \frac{1}{C}\sum_{i}\{\sum_{x\in A_i}[f(x) - \frac{1}{n_i}\sum_{x\in A_i} f(x)]^2\} \tag{23}$$

Where $A_i$ denotes different segmentation regions, $C$ is the normalization constant, $f(x)$ is the gray value of point $x$ in the image, $n_i$ is the number of the pixels in each region $A_i$. The closer to 1 the value of $pp$ is, the more uniform the interior of the segmentation regions are and the better the quality of the segmentation is.

**2). Dice similarity coefficient**

The ground truth (GT) is drawn manually through visual inspection of the SAR images. The Dice Similarity coefficient (DSC) [24] between the computed segmentation (CS) and the GT is defined as:

$$DSC(CS, GT) = 2 \times \frac{N(CS \cap GT)}{N(CS) + N(GT)} \tag{24}$$

Where $N(\cdot)$ indicates the number of voxels in the enclosed set. The closer the DSC value to 1, the better the segmentation.

## Ⅳ. EXPERIMENTS RESULTS

In this section, two synthetic images (by setting $L = 2$ for Gamma noise) are used to test the efficiency of our proposed model, whose size are $85 \times 76$ and $85\times 61$ , respectively. Two ERS-2 SAR images have the size $398\times 344$ and $240\times 279$ , respectively , and the gray-scale in the range between 0 and 255. All the algorithms are implemented with Matlab 8.0 in core2 with 1.9 GHZ and 1GB RAM.

For fair comparison with GID model , we give a version of the model described in [8] with edge detection function and a level set regularization term, denoted by GAA. The level set function of GAA and GID model all can be simply initialized as a binary step function which takes a constant value 1 inside a region and another constant value -1 outside.

For synthetic image1 and image2, we adopt the following parameters. The parameters of GAA model are chosen as $\mu=255$ , $\Delta t=0.1$ , $\varepsilon=1$ , $v=1$ .The parameters of GID model are chosen as $\mu=3.0$ , $\Delta t=0.1$ , $\varepsilon=1$ , $\sigma=1.2$ , $\beta=100$ , $v=1$ . The parameters of GAA+GO and GID+GO are all chosen as $\mu=5$ , $\lambda=0.01$ , , $\sigma=1.2$ , $\beta=100$ . The parameters of GAA+FPA and GID+FPA are all chosen as $\mu=5$ , $\lambda=1$ , $\alpha=10$ , $\sigma=1.2$ , $\beta=100$ , $t=1e-5$ . The thresholding values for GAA+GO, GAA+FPA, GID+ GO and GID+FPA are all chosen as $\gamma=0.5$ , which are used to find the segmented region $\Omega_1=\{x:\phi(x)>\gamma\}$ .

For synthetic image1 and image2, we compare the speed of algorithms ( the pair (.,.) is used to report both the number of iterations ( the first number ) and the cpu time ( the second number )) and $DSC$ , which are listed in TABLE Ⅰ.

We show the final contours of two synthetic images in Fig.1 and Fig.2. We can see from Fig.1 and Fig.2 that the proposed GID model can also automatically detect the exterior and interior boundaries of two synthetic images same as GAA model. We also observe that GAA+GO or GID+GO algorithm can reduce much time than GAA or GID algorithm based on level set method, and GAA+FPA or GID+FPA algorithm can further reduce about half of the running time needed for GAA+GO or GID+GO algorithm. As can be seen from TABLE Ⅰ, the $DSC$ values of GID model are higher than those of GAA model.

For SAR image1 and image2, we compare the speed of algorithms ( the pair (.,.) is used to report both the number of iterations ( the first number ) and the cpu time ( the second number )) and $pp$, which are listed in TABLE Ⅱ.

We show the final contours of two SAR images in Fig.3 and Fig.4. We can see from Fig.3 and Fig.4 that the proposed GID model can also automatically detect the exterior and interior boundaries of two SAR images with severe multiplicative gamma noise same as GAA model. We also observe that GAA+GO or GID+GO algorithm can reduce much time than GAA or GID algorithm based on level set method, and GAA+FPA or GID+FPA algorithm can further reduce about half of the running time needed for GAA+GO or GID+GO algorithm.

As can be seen from TABLE Ⅱ, the $pp$ values of GAA+GO,GAA+FPA, GID+ GO and GID+ FPA are all close to 1, which show high precision of the proposed fast algorithm.

## Ⅴ.CONCLUSION

In this paper, we propose a novel variational active contour model based on I-divergence-TV model, which hybrides edge-based model with region-based model and can be used to segment images corrupted by multiplicative gamma noise. We transform the proposed model into classic ROF model by adding a proximal operator. We propose a fast fixed point algorithm to solve SAR image segmentation question. Experimental results for synthetic images and real SAR images show that the proposed image segmentation model can efficiently stop the contours at weak or blurred edges, and can automatically detect the exterior and interior boundaries of images with multiplicative gamma noise. The proposed fast fixed point algorithm is robust to initialization contour, and can further reduce about 10% of the time needed for algorithm proposed by Goldstein-Osher.

The reason why the proposed algorithm is faster than the GO algorithm is that it do not involve partial differential or difference equations. The proposed algorithm can also be applied to isotropic TV-based model. We find that image segmentation model can be transformed into the ROF model, which has been intensively studied in literature.

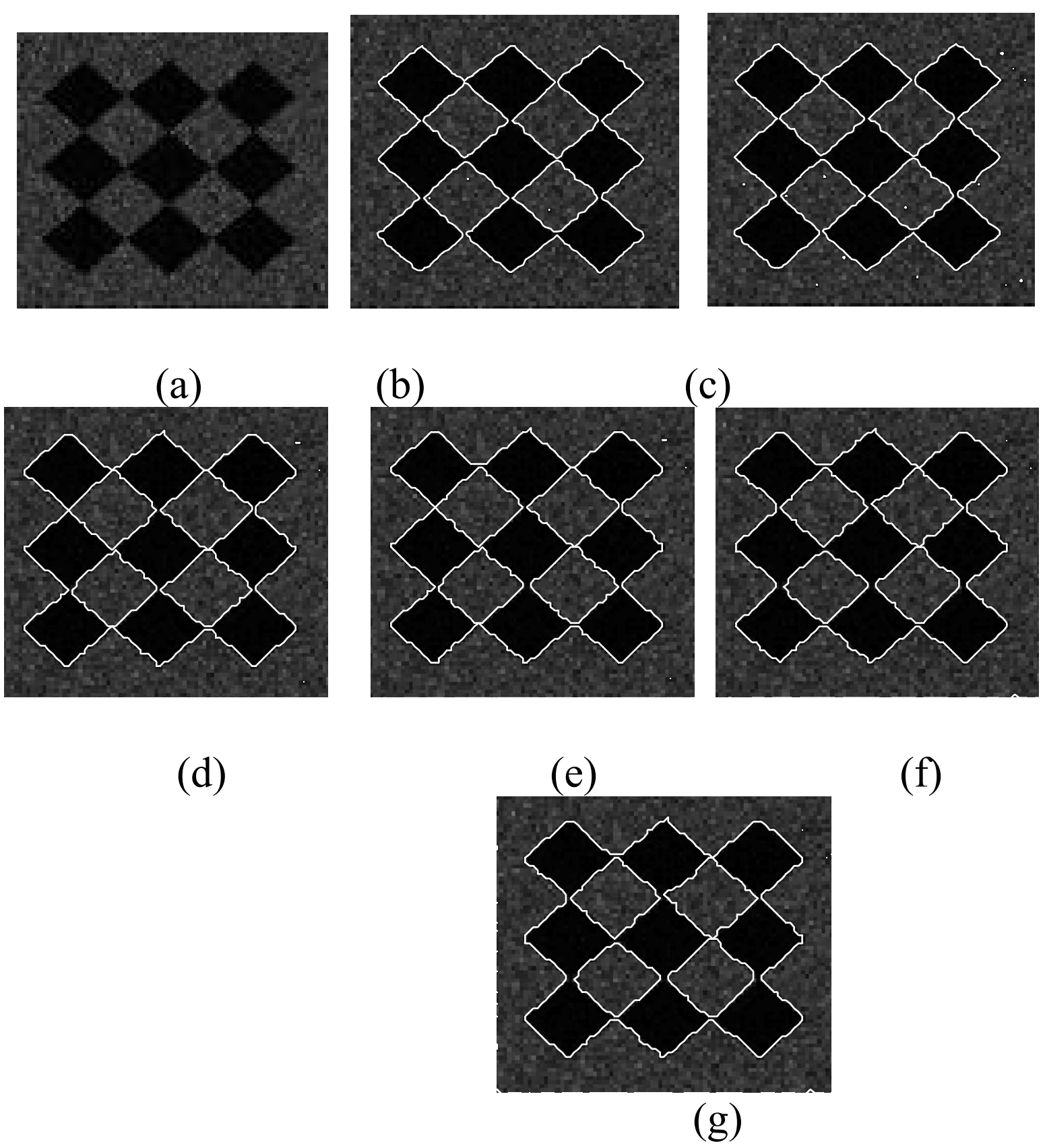

Fig.1. (a) synthetic image 1. (b) final contour by GAA.(c) final contour by GID. (d) final contour by GAA +GO. (e) final contour by GID +GO.(f) final contour by GAA+FPA. (g) final contour by GID +FPA.

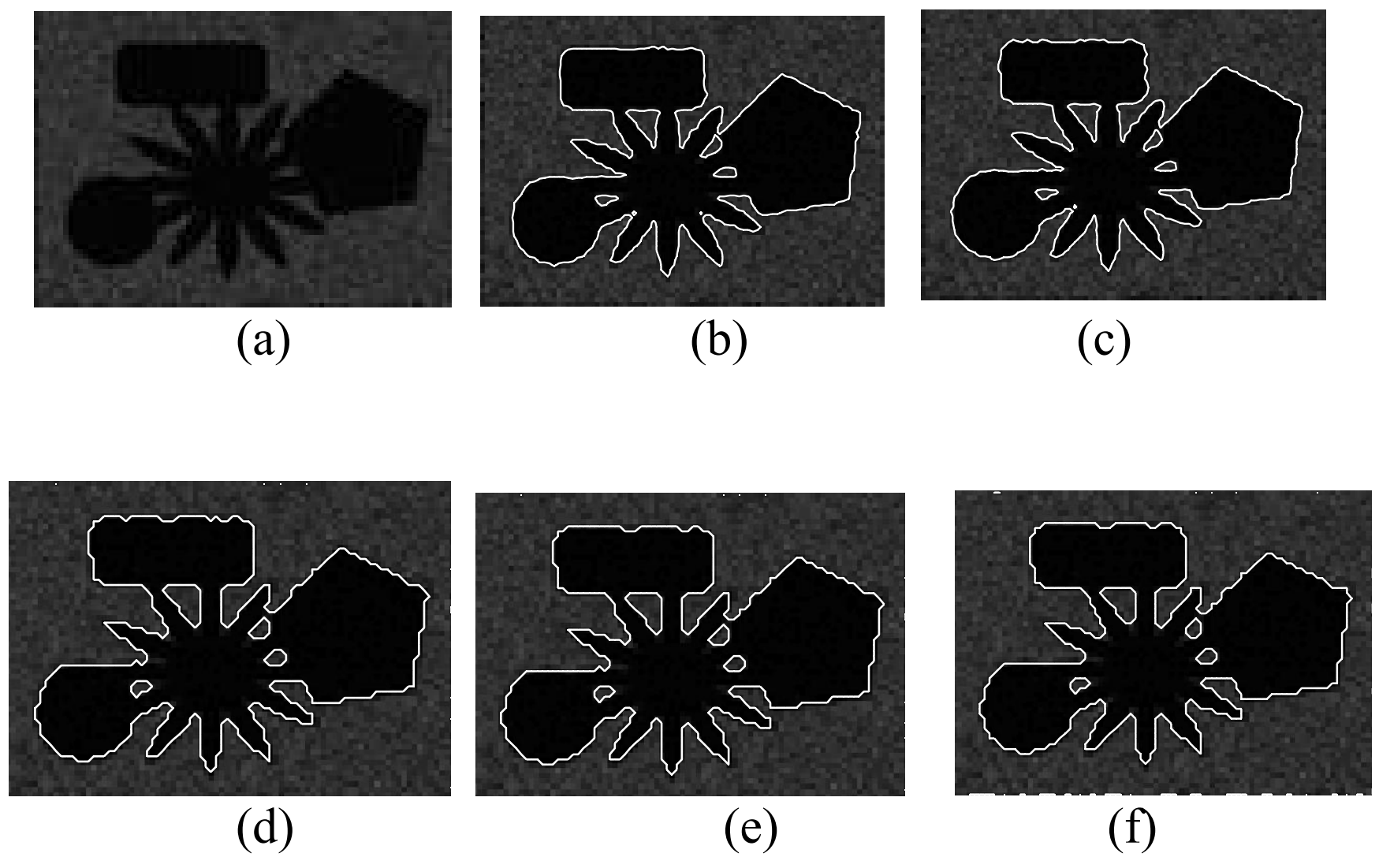

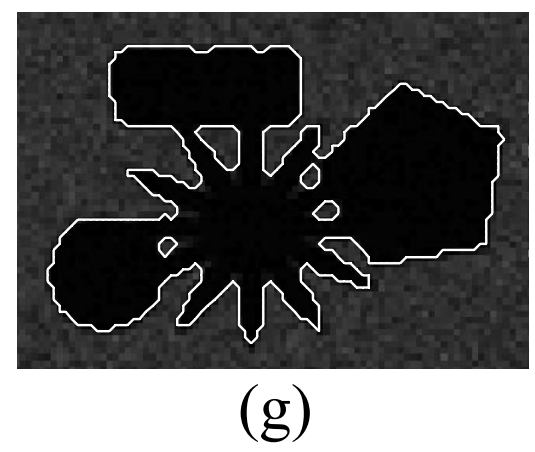

Fig.2. (a) synthetic image 2. (b) final contour by GAA.(c) final contour by GID. (d) final contour by GAA +GO. (e) final contour by GID +GO.(f) final contour by GAA+FPA. (g) final contour by GID +FPA.

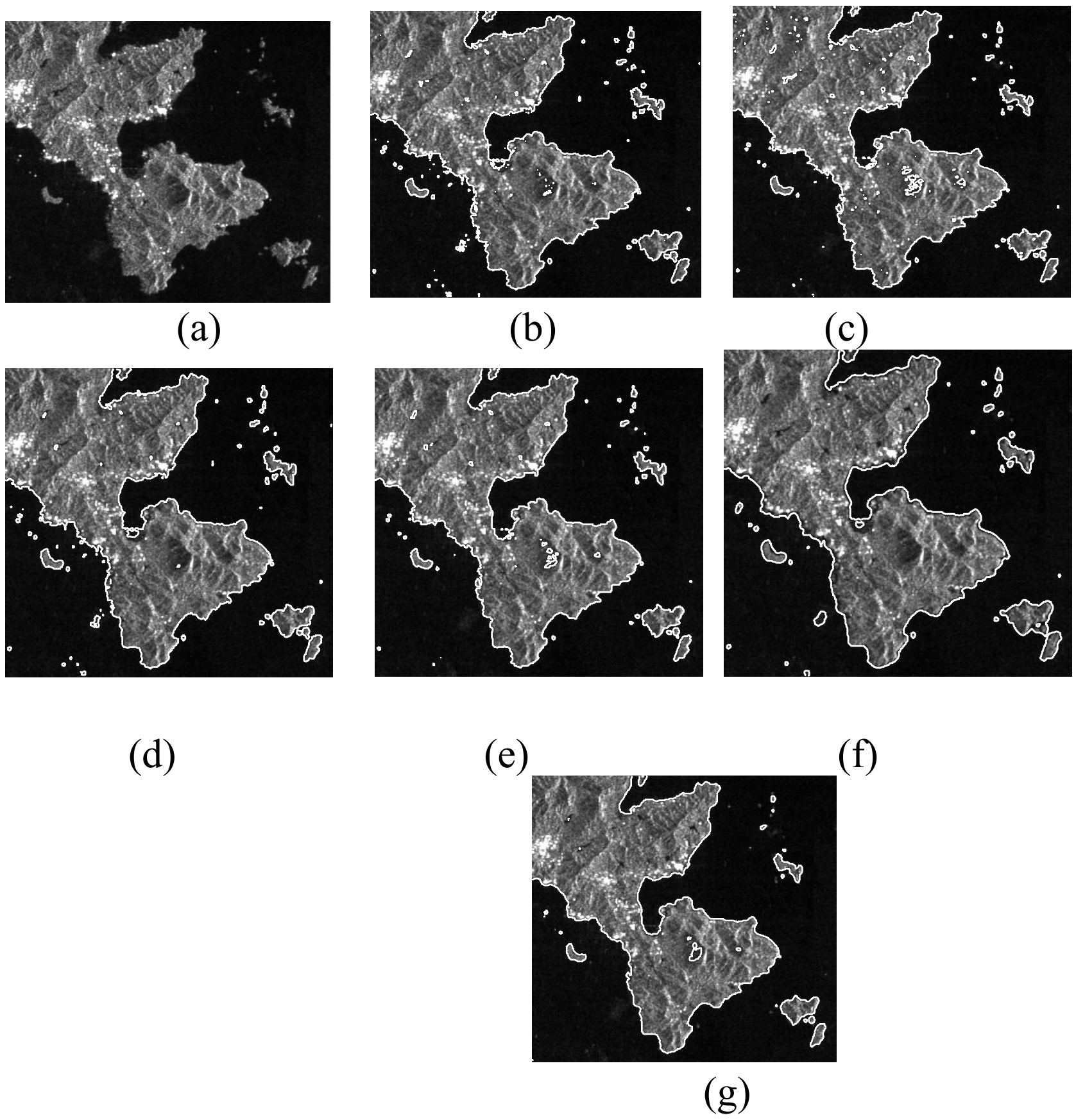

Fig.3. (a) SAR image1. (b) final contour by GAA.(c) final contour by GID. (d) final contour by GAA +GO. (e) final contour by GID +GO.(f) final contour by GAA+FPA. (g) final contour by GID +FPA.

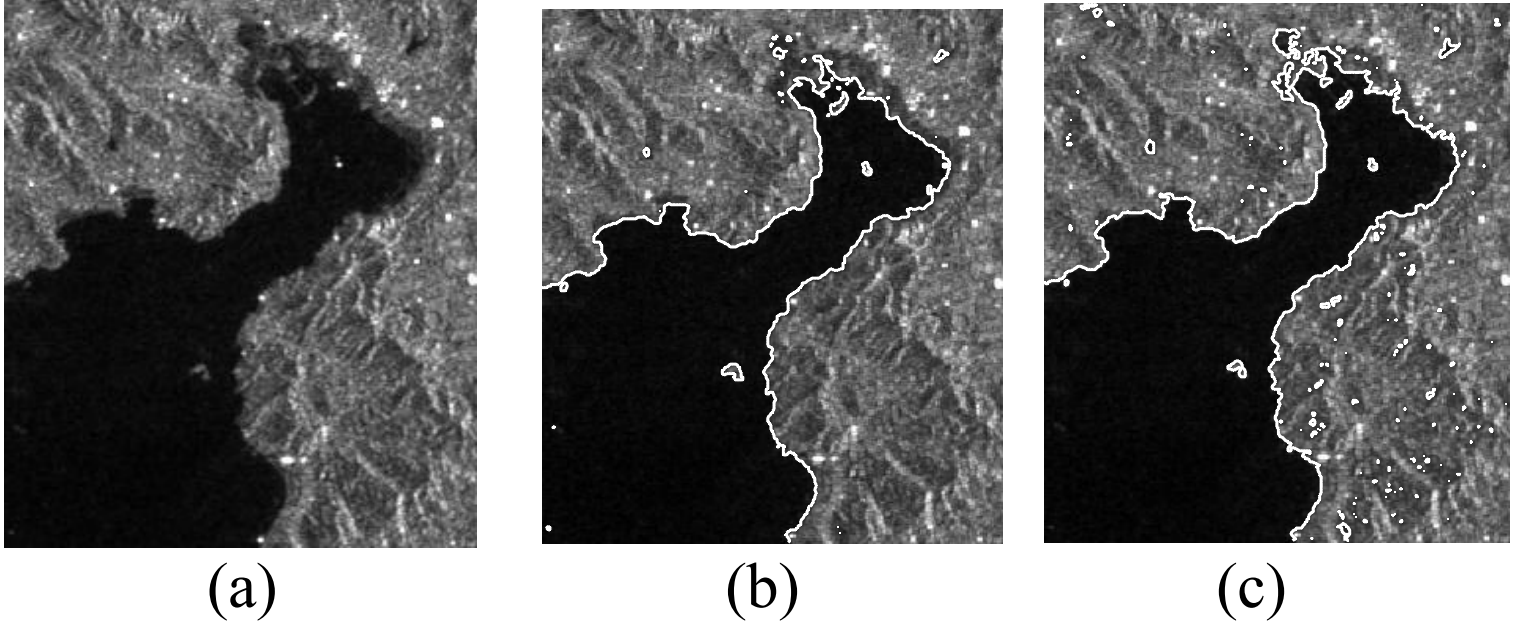

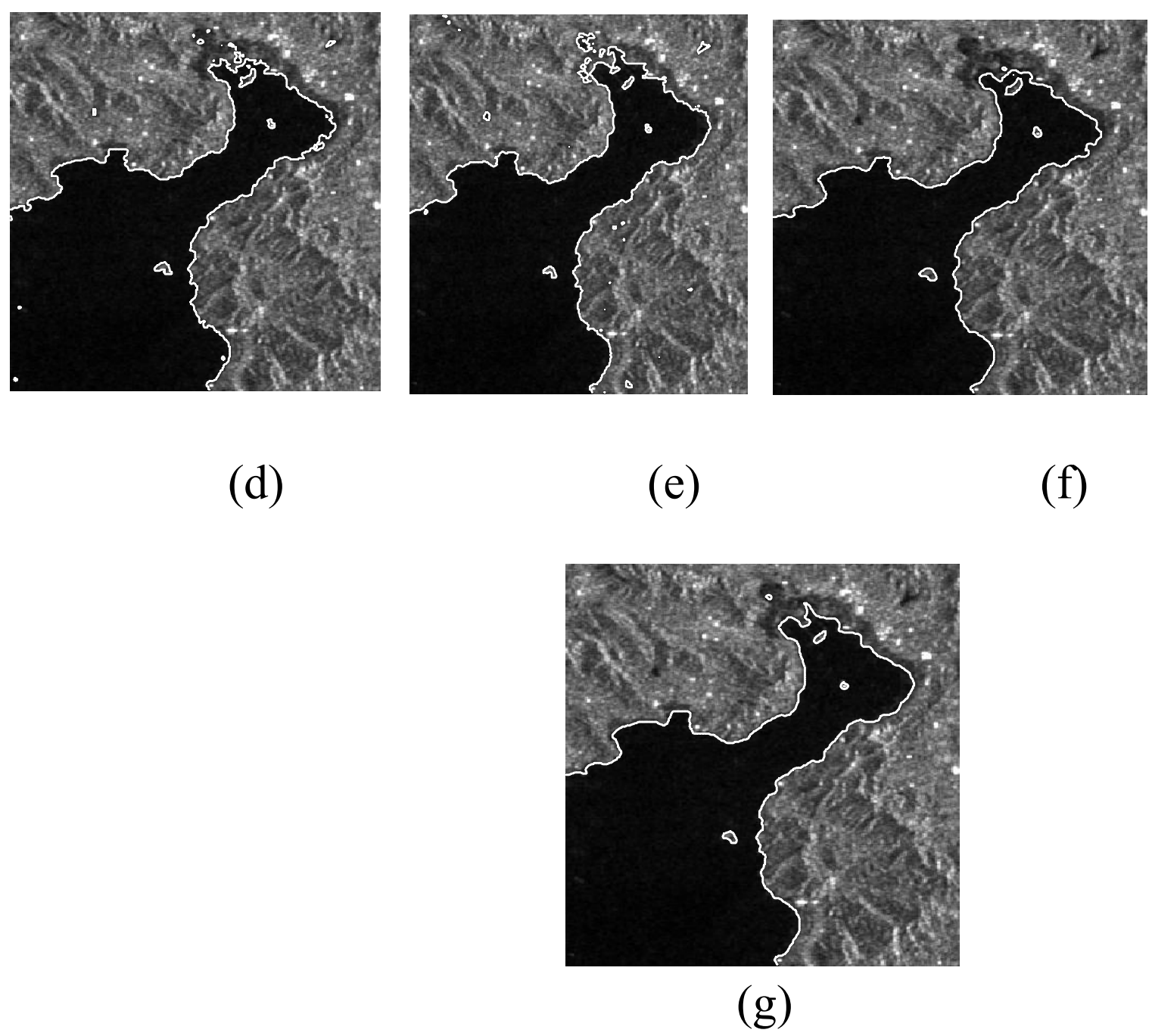

Fig.4. (a) SAR image2. (b) final contour by GAA.(c) final contour by GID. (d) final contour by GAA +GO. (e) final contour by GID +GO.(f) final contour by GAA+FPA. (g) final contour by GID +FPA.

TABLE Ⅰ

PERFORMANCE OF GAA AND GID ALGORITHM

| Method | synthetic | image1 | synthetic | image 2 |
|---|---|---|---|---|
| | Speed | DSC(PP) | Speed | DSC(PP) |
| GAA | (20,0.531s) | 97.37% (0.958) | (20,0.39s) | 97.55% (0.9295) |
| GID | (20,0.422s) | 98.58% (0.932) | (20,0.36s) | 99.11% (0.9133) |
| GAA+GO | (10,0.375s) | 96.84% (0.9993) | (10,0.344s) | 97.69% (0.9992) |
| GID+GO | (10,0.359s) | 96.70% (0.9993) | (10,0.344s) | 97.59% (0.9992) |
| GAA+FPA | (10,0.188s) | 95.88% (0.9993) | (10,0.176s) | 96.90% (0.9992) |
| GID+FPA | (10,0.175s) | 95.76% (0.9993) | (10,0.18s) | 96.78% (0.9992) |

TABLE Ⅱ

PERFORMANCE OF GID, GO+GID AND FPA+GID ALGORITHM

| Method | SAR image1 | | SAR image 2 | |
|---|---|---|---|---|
| | Speed | PP | Speed | PP |
| GAA | (50,17.907s ) | 0.9769 | (50,8.937) | 0.9588 |
| GID | (50,17.625s ) | 0.9875 | (50,8.734) | 0.9784 |
| GAA+GO | (30,9.238s ) | 0.999 | (30,4.875s ) | 0.999 |
| GID+GO | ( 30,9.046s ) | 0.999 | (30,4.753s ) | 0.999 |
| GAA+FPA | (30,4.362s ) | 0.999 | (30,2.439s ) | 0.999 |
| GID+FPA | (30,4.681s ) | 0.999 | (30,2.354s ) | 0.999 |